\documentclass[twocolumn,letterpaper]{IEEEAerospaceCLS}  
\pdfoutput=1

\usepackage[]{graphicx}
\usepackage{algorithm}
\usepackage{algpseudocode}
\usepackage{amsmath}
\usepackage{amssymb}
\usepackage{tikz}
\usepackage{pgfplots}
\usepackage{ifthen}  
\usepackage{xcolor}
\usepackage{dblfloatfix}
\PassOptionsToPackage{hyphens}{url}\usepackage[hidelinks]{hyperref}
\usepackage[capitalize]{cleveref}
\usepackage{multirow}
\normalsize
\usepackage{booktabs}
\usepackage{csquotes}

\usepackage{tablefootnote}
\usepackage{siunitx}

\usepackage[numbers]{natbib}

\newcommand{\ignore}[1]{}

\begin{document}
\title{Markov Decision Processes for Satellite Maneuver Planning and Collision Avoidance}

\author{
William Kuhl, Jun Wang, Duncan Eddy, Mykel J. Kochenderfer\\ 
Stanford University\\
496 Lomita Mall\\
Durand Building\\
Stanford, California 94305\\
{billkuhl}, {jun2026}, {deddy}, {mykel}@stanford.edu
\thanks{\footnotesize 979-8-3503-5597-0/25/$\$31.00$ \copyright2025 IEEE}
}

\maketitle

\thispagestyle{plain}
\pagestyle{plain}

\maketitle

\thispagestyle{plain}
\pagestyle{plain}

\begin{abstract}
This paper presents a decentralized, online planning approach for scalable maneuver planning for large constellations.
While decentralized, rule-based strategies have facilitated efficient scaling, optimal decision-making algorithms for satellite maneuvers remain underexplored. 
As commercial satellite constellations grow, there are benefits of online maneuver planning, such as using real-time trajectory predictions to improve state knowledge, thereby reducing maneuver frequency and conserving fuel. 
We address this gap in the research by treating the satellite maneuver planning problem as a Markov decision process (MDP).
This approach enables the generation of optimal maneuver policies online with low computational cost.
This formulation is applied to the low Earth orbit collision avoidance problem, considering the problem of an active spacecraft deciding to maneuver to avoid a non-maneuverable object.
We test the policies we generate in a simulated low Earth orbit environment, and compare the results to traditional rule-based collision avoidance techniques.

\end{abstract}

\tableofcontents

\section{Introduction}

The number of satellites in low earth orbit (LEO) has increased as corporations gain access to affordable space launch. 
This increase in constellations and other missions heightens the potential for collisions between active satellites and non-maneuverable objects. 
To mitigate this risk, satellite constellation operators execute maneuvers frequently.  
For instance, in a six-month period SpaceX's starlink constelation executed over 25,000 maneuvers in 2021 and 50,000 in 2024~\cite{pultarova_performing_2023,pultarova_spacex_2024}.

Current maneuver planning methods use fixed rule-based policies or offline methods that do not sufficiently account for potential information updates, often leading to suboptimal outcomes. 
This paper demonstrates how optimized decision-making algorithms can enhance the planning of collision avoidance maneuvers. 
It specifically addresses the problem of how far in advance a satellite operator should maneuver to minimize fuel cost and probability of collision.

Traditionally, collision avoidance maneuvers are planned using algorithms run on terrestrial computers, with plans uploaded to spacecraft in advance of projected collisions~\cite{hawkins_flight_2017}. 
As communication with satellites in LEO approaches real-time, operators can wait to receive more information updates before planning a maneuver.  
Recently, some constellation operators have switched to online collision avoidance algorithms that run continuously on satellites~\cite{hejduk2023conjunction}.
Whether computed on-orbit or terrestrially, current algorithms use rule-based policies to decide if and when a maneuver is necessary.

\begin{figure}
    \includegraphics[width=\linewidth]{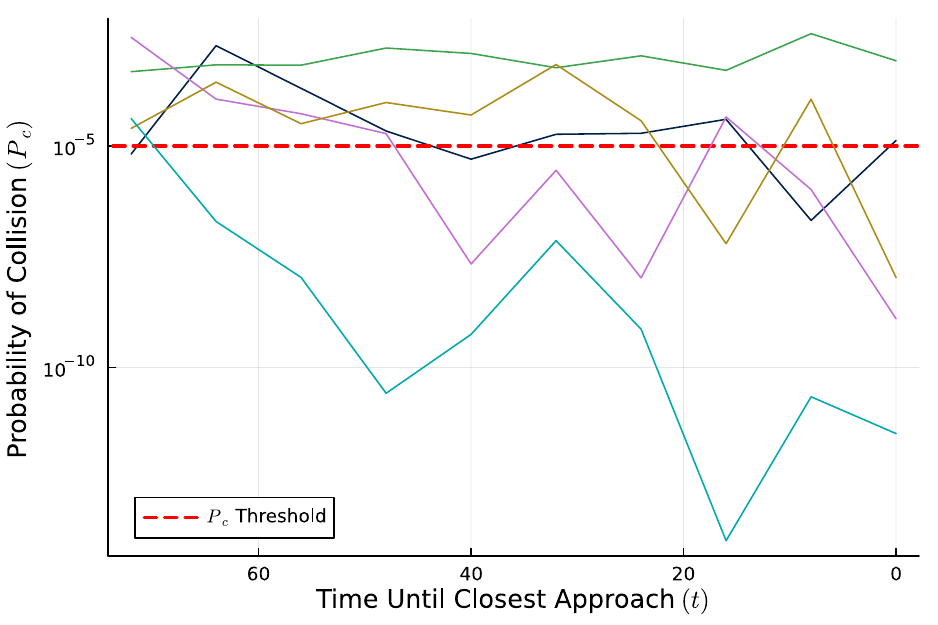}
    \caption{Simulated encounters between a satellite and piece of debris. The line at $P_c=\num{e-5}$ represents a common operator defined risk threshold. As the time until closest approach decreases $P_c$ is updated to reflect new state estimates.}
    \label{fig:encounterex}
\end{figure}

When a potential collision is imminent, the U.S. Space Force's 18th or 19th Space Defense Squadron will issue a Conjunction Data Message (CDM) to the satellite operators as a source of information to help them plan maneuvers to reduce their probability of collision. 
Based on the information in a CDM and any additional knowledge of their spacecraft's state, operators decide how to mitigate collision risk. 
The probability of collision can fluctuate in CDM updates as the time of closest approach nears. 
This requires operators to weigh the benefits of early maneuvers, which may incur lower fuel costs, against the need for more accurate data. 
While the probability of collision for many encounters ultimately falls below a risk threshold after state updates, waiting for these updates sometimes results in a satellite executing a late maneuver with greater fuel expenditure.
An example of how the probability of collision $P_c$ changes as the time until closest approach $t$ decreases is shown in~\cref{fig:encounterex}.

There has been extensive work in spacecraft maneuver planning for close proximity operations.
\citeauthor{silvestrini_neural-based_2021} use a model-based reinforcement learning approach with inverse reinforcement learning and long short-term memory algorithms for the guidance and control of distributed formation flying missions~\cite{silvestrini_neural-based_2021}.
\citeauthor{bevilacqua_development_2011} outline the use of linear quadratic regulators and artificial potential functions for autonomous proximity maneuvering~\cite{bevilacqua_development_2011}. 

Recently, there have been several studies into online and on-board methods for autonomous collision avoidance. 
SpaceX and NASA present an experiment to test the close approach encounters between the Starlink constellation and NASA's Starling formation~\cite{hejduk2023conjunction}. 
\citeauthor{slater_collision_2006} detail collision avoidance methods for formation flight in LEO~\cite{slater_collision_2006}. \citeauthor{stoll_operational_2013} discuss the impact of external satellite and debris state information sources for the operations of small satellites~\cite{stoll_operational_2013}. 

Markov decision process and reinforcement learning methods have been presented to solve the Earth-observing satellite task planning problem.
\citeauthor{eddy_markov_2019} apply decision-making techniques in the context of image collection planning~\cite{eddy_markov_2019}. 
\citeauthor{herrmann_monte_2022} apply reinforcement learning and Monte Carlo methods also in the context of the Earth-observing satellite scheduling problem~\cite{herrmann_monte_2022, herrmann_reinforcement_2023}.

This paper specifically investigates how far in advance of a projected conjunction a satellite operator should  execute an evasive maneuver to minimize both fuel cost and collision probability given an incoming series of information updates.
We propose multiple methods for solving this problem within the framework of Markov decision processes (MDPs). 
An MDP formulation enables the use of powerful decision-making tools. %
This paper analyzes how these methods perform when tasked with mitigating risk of a potential collision between a satellite and a non-maneuverable object. 
This paper's focus is on practical methods for satellite collision avoidance that only depend on the information in CDMs.

\section{Markov Decision Process Formulation}

A Markov decision process (MDP) is a general framework for solving decision-making problems. An agent at a state chooses an action, receives a reward, and transitions to the next state with some probability~\cite{kochenderfer_algorithms_2022}.
An MDP is defined by a state space $\mathcal{S}$, an action space $\mathcal{A}$, a transition function ${T}$, and a reward function ${R}$:
\begin{equation}
    {\mathcal M} = ({\mathcal{S}},{\mathcal{A}},T,R)
\end{equation}

This paper examines the case where an operator-controlled satellite (the chief satellite), approaches a piece of unpowered debris (the deputy satellite). 
There is uncertainty in the chief and deputy state at the time of closest approach. 
As the time of closest approach nears, state information is periodically updated with increasing precision. 

\subsection{State Space}

We define the state space $\mathcal{S}$ as the information an operator would receive in a typical CDM
\begin{equation}
    \label{eq:state_def}
    s = (t,x_c,x_d,\Sigma_c,\Sigma_d,r_c,r_d)
\end{equation}
The variable $t\in[0,72]$ is the time (in hours) until the closest approach between the chief and deputy. 
The state contains the chief's Earth centered inertial (ECI) position and velocity $x_c$, the deputy's ECI position and velocity coordinates $x_d$, and respective covariance matrices $\Sigma_c$ and $\Sigma_d$. 
The hardbody radii of the chief $r_c$ and deputy $r_d$ are also included in the state variable.

The chief and deputy ECI Coordinates are vectors of the satellite's three-dimensional position and velocity defined as
\begin{equation}
    \label{eq:eci_def}
    x = (x,y,z,\dot x, \dot y,\dot z)
\end{equation}

\subsection{Action Space}
The action space is ${\mathcal{A}} = \{\text{wait}, \text{maneuver}\}$.
At each discrete step, the chief satellite $x_c$ chooses from the set of feasible actions $\mathcal{A}$. 
For every case where $t > 0$, the chief must decide whether to execute an evasive maneuver or to wait for another state update. 
We limit the evasive maneuver thrusts to along-track and anti-along track, and execute the evasive maneuver according to the policy in \cref{appendix:cost2maneuver}. 

When $t = 0$ the satellites have already crashed or not, no action is necessary.

\subsection{Transition Function}
The transition function $T(s,a)$ defines a distribution from which the next state is sampled. 
This distribution encodes how we simulate potential state updates over time. 
As previously discussed, the chief is actively controlled and tracked frequently, while the deputy is non-maneuverable and tracked sporadically. 
Hyperparameter $p_\text{obs}$ represents the probability the deputy is observed.
If the deputy is observed, the deputy state $x_d$ is updated and the deputy covariance $\Sigma_d$ is scaled by $(1-k_d)$. The scalar $k_d$ is sampled from a uniform distribution over the range $\left[k_{d,\text{lower}},k_{d,\text{upper}}\right]$.
The chief's covariance is scaled by $(1-k_c)$ every state update. 
This behavior is defined by the equations: 
\begin{align}
    t &\leftarrow t - 8 \label{time_update}\\
    x_c &\sim {\mathcal{N}}(x_c, \Sigma_c) \label{chief_state_update} \\
    x_d &\sim {\mathcal{N}}(x_d, \Sigma_d) \label{deputy_state_update} \\
    \Sigma_c &\leftarrow \Sigma_c (1-k_c) \label{chief_covariance_update} \\
    k_d &\sim {\mathcal{U}}(k_{d,\text{lower}},k_{d,\text{upper}}) \label{deputy_cov_param_sample} \\
    \Sigma_d &\leftarrow
    \begin{cases} 
        \Sigma_d (1-k_d) & \text{with prob. } p_\text{obs}, \\
        \Sigma_d & \text{otherwise}.
    \end{cases} \label{deputy_covariance_update}
\end{align}
\cref{time_update} advances the time until closest approach by 8 hours. 
\cref{chief_state_update} updates the chief state $x_c$ by sampling the normal distribution defined by $x_c$ and $\Sigma_c$. 
\cref{deputy_state_update} similarly updates $x_d$ by sampling the normal distribution defined by $x_d$ and $\Sigma_d$. 
\cref{chief_covariance_update} updates the next agent covariance state $\Sigma_c$ by decreasing it by factor $k_c$. 
\cref{deputy_covariance_update} updates the deputy covariance with probability $p_\text{obs}$. 
The hardbody radii $r_c$ and $r_d$ remain unchanged.

\subsection{Reward Function}

The reward function $R(s,a)$ defines the reward (if positive) or penalty (if negative) for a given state and action.
We design the reward function to penalize the agent for expending fuel to maneuver or assuming too much risk.
These negative rewards are defined by the change in velocity $\Delta v$ required to mitigate the probability of collision $P_c$. 
The agent also receives a large penalty if $P_c$ is above a threshold $P_{c,\text{threshold}}$ at $t=0$ associated with a failure to properly mitigate the collision risk.

The cost to maneuver $R_\text{maneuver}$ is the minimum $\Delta v$ required at the time a CDM is received $t_\text{CDM}$ to ensure $P_c<P_{c,\text{threshold}}$ at $t=0$ .
Because $x_c$ is the chief position at $t=0$, we backpropagate it to $t_\text{CDM}$.
Thrust $\Delta v$ is then applied such that $P_c<P_{c,\text{threshold}}$ 
 when $x_c$ is propagated forward to $t=0$.
The total $\Delta v$ used is the cost to maneuver.
\cref{appendix:cost2maneuver} describes the specific algorithm used to calculate this cost. 
The probability of collision is calculated based on the method outlined by Foster~\cite{foster_parametric_1992}. The reward policy is outlined in \cref{rewards}.

\begin{table}[ht]
\caption{\bf MDP State-Action Rewards}
\centering
\begin{tabular}{l r}
\toprule
\bfseries Reward & \bfseries Condition \\
\midrule

$R_{maneuver}$   & $t \neq 0$ and $a = \text{maneuver}$ \\
$R_{crash}$ & $t = 0$ and $P_c > P_{c,\text{threshold}}$ \\
0     & otherwise \\

\bottomrule
\end{tabular}
\label{rewards}
\end{table}

\subsection{Solution Methods}
To solve this problem, we use Monte Carlo tree search (MCTS)~\cite{kochenderfer_algorithms_2022}.
MCTS is an online method that calculates the optimal action from a current state by exploring the set of reachable future states.
MCTS extracts the optimal action from the state-action space $Q(s,a)$ with state $s\in S$ and action $a \in \mathcal A$.
The optimal action $a_\text{best}$ is the action that maxizes $Q$ at state $s$ 
\begin{equation}
a_\text{best} = \text{argmax}_a Q(s,a)
\end{equation}

In MCTS, the state-action space is explored by sampling actions from known states. 
When an action is sampled, MCTS samples the next state $s'$ from $T(s,a)$. 
If $s'$ has not been visited before, it is added to the set of visited states $V$ and returns 0, otherwise a new action $a'$ is sampled at the new state.
The reward $R(s,a)$ plus the the returned value from explored states discounted by $\gamma \in [0,1]$ is used to update the state-action value function $Q(s,a)$.
Each time a state-action pair is sampled, a count variable $N(s,a)$ is incremented by one.  
This process of sampling is executed $N_\text{sim}^\text{max}$ times, after which the action that maximizes $Q(s,a)$ is returned. 

Due to the high degree of uncertainty in the transition function and the continuous state space, it is extremely improbable that a state will be sampled more than once.
This results in a search tree that never reaches a depth of more than one timestep ahead which is not sufficient to model future risk.

To address this issue, we modify the MCTS algorithm by replacing $Q(s,a)$ with a time-action value function $Q_t(t,a)$ and a corresponding time-action count $N_t(t,a)$.
Because transitions between the time element of the state are deterministic, MCTS using $Q_t$ will re-visit time states frequently and will sample multiple timesteps in the future. 
\cref{mcts_cp} provides this modified implementation of MCTS.

\begin{algorithm}
    \caption{Monte Carlo Tree Search}
    \begin{algorithmic}[1] 
       \Function{Simulate}{$s,V,N_t,Q_t$}
            \State $t, x_c,x_d, \Sigma_c, \Sigma_d, r_c, r_d \leftarrow s$
            \If {($t$,$a$) $\notin V~\forall~a \in {\mathcal{A}}$}
                \For {$a \in \mathcal{A}$}
                    \State $N_t(t,a)\leftarrow 0 $
                    \State $Q_t(t,a)\leftarrow 0.0$
                    \State $V \leftarrow V \cup \{(t,a)\}$
                \EndFor
                \State \Return $0,N_t,Q_t$
            \EndIf
            \If {$t = 0$} \label{mcts:edit1}
                \State \Return $R(s,a)$
            \EndIf
            \State $a \leftarrow $ \Call{explore}{s} \label{line:explore}
            \State $s' \leftarrow $ $T(s,a)$
            \State $r \leftarrow $ $R(s,a)$
            \State $q \leftarrow r + \gamma$\Call{Simulate}{$s',V,N_t,Q_t$}
            \State $N_t(t,a) \leftarrow N_t(t,a) + 1$
            \State $Q_t(t,a) \leftarrow Q_t(t,a) + \left( q - \frac{q-Q_t(t,a)}{N_t(t,a)}\right)$
            \State \Return $q, N_t, Q_t$
        \EndFunction
        \Function{MCTS}{$s_0$,$\gamma$}
        \State $t, x_c,x_d, \Sigma_c, \Sigma_d, r_c, r_d \leftarrow s_0$
        \State $V,N_t,Q_t \leftarrow \emptyset, \emptyset, \emptyset$
            \For {$n \leq N_\text{sim}^\text{max}$}
                \State $q, N_t, Q_t \leftarrow \text{\Call{Simulate}{$s_0,V,N_t,Q_t$}}$
                \State $n \leftarrow n + 1$
            \EndFor
            \State \Return $a \leftarrow \text{argmax}_a Q_t(t,a)$
        \EndFunction
        \Function{RunSequence}{$s$}
            \State $t, x_c,x_d, \Sigma_c, \Sigma_d, r_c, r_d \leftarrow s$
            \While {$t > 0$} \label{mcts:edit2}
                \State $a \leftarrow$ \Call{MCTS}{$s$} 
                \If {$a$ = maneuver}
                    \State $R_\text{maneuver} \leftarrow R(s,a)$
                    \State \Return $R_\text{maneuver}$, $s$, $a$
                \Else
                    \State s = \Call{RecieveStateUpdate}{$s$}
                \EndIf
            \EndWhile
            \If{$P_c(s) > P_{c,\text{threshold}}$}
                \State \Return $R_\text{crash}$, $s$, wait
            \Else      
                \State \Return 0, $s$, wait
            \EndIf
        \EndFunction
    \end{algorithmic}
    \label{mcts_cp}
\end{algorithm}

\subsubsection{MCTS: Limited Horizon}

As the cost for maneuvering $R_\text{maneuver}$ is determined by orbital dynamics, the hyperparameter $R_\text{crash}$ determines how much risk the chief is willing to accept.  
If $R_\text{crash}$ is too large, the chief maneuvers early whenever a potential collision is sampled, no matter how improbable.
If this penalty is too small, however, there is a greater risk a collision will take place. 
A suitable $R_\text{crash}$ can be determined by conducting a search over simulated close encounters. 
However, without algorithmic guarantees of safety many trials are required to verify the risk assumed by deploying MCTS is within acceptable bounds.

We propose a modification to \cref{mcts_cp} which guarantees the same level of risk as rule-based planners while only considering penalties that come from system dynamics. 
A floor on risk is set by limiting the planning horizon to one timestep before the time of closest approach.
This is achieved by replacing \enquote{0} in \cref{mcts:edit1} and \cref{mcts:edit2} with $t_\text{maxdepth}$.
The final reward is then the cost of maneuvering at $t_\text{maxdepth}$ if $P_c<P_{c,\text{threshold}}$ and 0 otherwise. 

\subsection{Exploration Policies}

A key component of MCTS is the method used to select the next action when sampling future rewards.
Because MCTS uses a limited number of simulation, a heuristic policy is required to select the next action to test.
In \cref{mcts_cp} this is represented as the function \enquote{explore} in \cref{line:explore}.
As a part of testing MCTS with a full horizon and limited horizon, we incorporate two different exploration heuristics: UCB1 and stochastic depth.

\subsubsection{UCB1}
A common heuristic for exploring the state-action space is the UCB1 algorithm~\cite{kochenderfer_algorithms_2022}.
Detailed in \cref{ucb1}, UCB1 uses the exploration constant $c$ along with the number of times a state action pair has been visited and the utility of that pair. This balances exploration of unexplored areas with the exploration of areas with the best utility. This works very well in a model with a large state-space, and is often the default exploration policy for MCTS. 

\begin{algorithm}
    
    \caption{Exploration Heuristic: UCB1}
    \begin{algorithmic}[1] 
        \Function{bonus}{$N_t,t,a$}
            \If {$N_t(t,a) = 0$}
                \State \Return $\infty$
            \Else
                \State $N_{t,\text{sum}} \leftarrow \sum_a N_t(t,a)$
                \State \Return $\sqrt{\frac{\log{N_{t,\text{sum}}}}{N_t(t,a)}}$
            \EndIf
        \EndFunction
        \Function{explore}{$s$}
            \State $t, x_c,x_d, \Sigma_c, \Sigma_d, r_c, r_d \leftarrow s$
            \State \Return $\text{argmax}_a(Q_t(t,a) + c\times{\text{\Call{bonus}{$N_t,t,a$}}} )$
        \EndFunction
    \end{algorithmic}
    \label{ucb1}
\end{algorithm}

\subsubsection{Stochastic Depth Heuristic}

Because future states have greater uncertainty, there is increased variance in their potential values. 
Because of this, more information is gained by sampling states far in the future than states close to the current satellite time.
Algorithms like UCB1 may oversample early time-action pairs, wasting time on uninformative simulations.
In the time-action space, it is possible to define a stochastic policy that emphasises sampling future states that have higher reward variance.
\cref{stochastic_depth} outlines an example policy that samples a future state more often than a current action according to exploration parameter $c$.
\begin{algorithm}
    \caption{Exploration Heuristic: Stochastic Depth}
    \begin{algorithmic}[1] 
        \Function{explore}{$s$}
            \State $t, x_c,x_d, \Sigma_c, \Sigma_d, r_c, r_d \leftarrow s$
            \State $N_{t,\text{sum}} \leftarrow \sum_a N_t(t,a)$
            \If {$N_{t,\text{sum}}=0$}
                \State \Return \Call{Random}{$\mathcal A$}
            \ElsIf {$\frac{{N}_t(t,\text{wait})}{{N}_{t,\text{sum}}}>c$}
                \State \Return maneuver
            \Else
                \State \Return wait
            \EndIf
        \EndFunction
    \end{algorithmic}
    \label{stochastic_depth}
\end{algorithm}

\subsection{Baseline Methods}

We establish a set of static rules as a baseline for our maneuver policies, reflecting a fixed decision policy put in place by a satellite operator. 
A cutoff time $t_\text{cutoff}$ is set where $t_\text{cutoff}>0$. 
Until $t_\text{cutoff}$, the chief does not perform a maneuver. 
After $t_\text{cutoff}$, the chief maneuvers at the next time that it calls below the danger threshold.
\cref{rule_based_planner} outlines this policy.
\begin{algorithm}
    \caption{Rule-Based Planner}
    \begin{algorithmic}[1] 
        \Function {RulePlanner} {$s_0, t_\text{cutoff}$}
            \State $t, x_c,x_d, \Sigma_c, \Sigma_d, r_c, r_d \leftarrow s$
            \If{$t>0$}
                \If{$t_s \leq t_\text{cutoff}$ and $P_c(s) > P_{c,\text{threshold}}$}
                    \State \Return maneuver
                \Else
                    \State \Return wait
                \EndIf
            \Else
                \State \Return wait
            \EndIf
        \EndFunction
        \Function{RunSequence}{$s, t_\text{cutoff}$}
            \State $t, x_c,x_d, \Sigma_c, \Sigma_d, r_c, r_d \leftarrow s$
            \While {$t > 0$}
                \State $a \leftarrow$ \Call{RulePlanner}{$s,t_\text{cutoff}$} 
                \If {$a$ = maneuver}
                    \State $R_\text{maneuver} \leftarrow R(s,a)$
                    \State \Return $R_\text{maneuver}$, $s$, $a$
                \Else
                    \State s = \Call{RecieveStateUpdate}{$s$}
                \EndIf
            \EndWhile
            \If{$P_c(s) > P_{c,\text{threshold}}$}
                \State \Return $R_\text{crash}$, $s$, wait
            \Else      
                \State \Return 0, $s$, wait
            \EndIf
        \EndFunction
    \end{algorithmic}
    \label{rule_based_planner}
\end{algorithm}

\section{Experiments}

We evaluate the performance of these solution methods on simulated encounters in Low Earth Orbit.
An encounter is defined as a scenario where a CDM is received before $t = 0$ where $P_c \geq P_{c,\text{threshold}}$.
We set $P_{c,\text{threshold}}=\num{e-5}$ as the desired level of risk mitigation.
Performance is defined by two metrics: the cost in $\Delta v$ per encounter and the proportion of unsafe encounters where risk is successfully mitigated. 
The cost per encounter is the amount of $\Delta v$ required at some time before a close approach to reduce $P_c$ at $t=0$ below $P_{c,\text{threshold}}$.
An encounter is considered mitigated if as $t$ approaches 0 the $P_c$ drops below $P_{c,\text{threshold}}$. 

We test the two MCTS methods, full horizon and limited horizon, with both exploration heuristics, UCB1 and stochastic depth.
We also test nine baseline rule-based methods with different values for $t_\text{cutoff}$.

To evaluate these methods, we generate a set of synthetic encounters when $t \leq 72$ hours before the time of closest approach.
The frequency and timing of state updates are modelled after the frequency an operator would receive a CDM.
A CDM is received at $t=72$ hours with a new CDM issued every eight hours until $t=0$.
We model this as receiving one state update every 8 hours. 

The simulation considers conjunctions with chiefs in LEO with orbital parameters in the ranges listed in in \cref{chief_osc}.
A deputy is randomly generated near the chief subject per the rules in \cref{appendix:scenario_gen}. 
\begin{table}
\caption{\bf Orbital Parameter Ranges for Conjunction Generation}
\centering
\begin{tabular}{l r}
\toprule
\bfseries Orbital Parameter & \bfseries Range \\
\midrule
$a$   & [$R_\text{Earth} + 400\text{km}$, $R_\text{Earth} + 600\text{km}$] \\
$e$   & [$0.01$, $0.11$] \\
$i$   & [$75^\circ$, $90^\circ$] \\
$\Omega$   & [$45^\circ$, $90^\circ$] \\
$\omega$   & [$30^\circ$, $60^\circ$] \\
$M$   & [$0^\circ$, $360^\circ$]  \\
\bottomrule
\end{tabular}
\label{chief_osc}
\end{table}
We assume that the position uncertainty for both the chief and deputy is a 3D Gaussian, with greater uncertainty in the along-track direction compared to the cross-track and radial directions.
Although there is greater variance in the position uncertainty of objects in LEO, we model the magnitude of the chief and deputy state uncertainties on the example found in the \textit{Conjunction Summary Message Guide}~\cite{noauthor_conjunction_nodate}. 
These covariances are backpropagated stochastically according to the rules set in \cref{appendix:scenario_gen}.
Synthetic chief and deputy measurements are sampled from these uncertainties and the true states, simulating noisy measurements.
These synthetic measurements are used to represent the updates a satellite would get as the time of closest approach nears.
The hardbody radii for the chief and deputy are sampled uniformly from the range [$5$m, $10$m].

\subsection{Simulation Characteristics}

\begin{figure}
    \includegraphics[width=\linewidth]{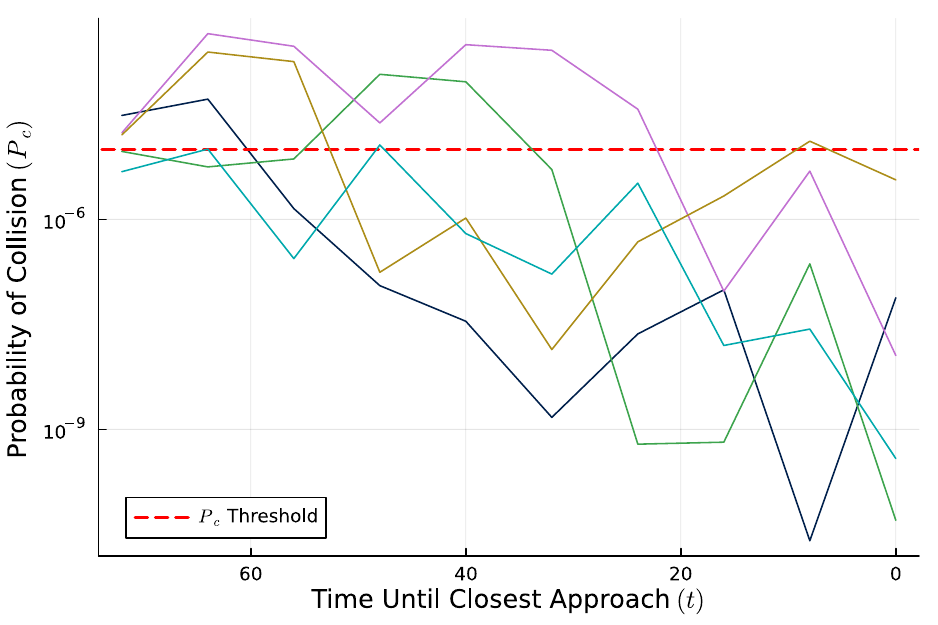}
    \caption{Examples of safe encounters. This plot shows how the probability of collision is updated as the time until closest approach decreases if no maneuver is taken. A safe encounter is defined as when the probability of collision falls below some threshold as the time until closest approach goes to 0.}
    \label{fig:safeenc}
\end{figure}

\begin{figure}
    \includegraphics[width=\linewidth]{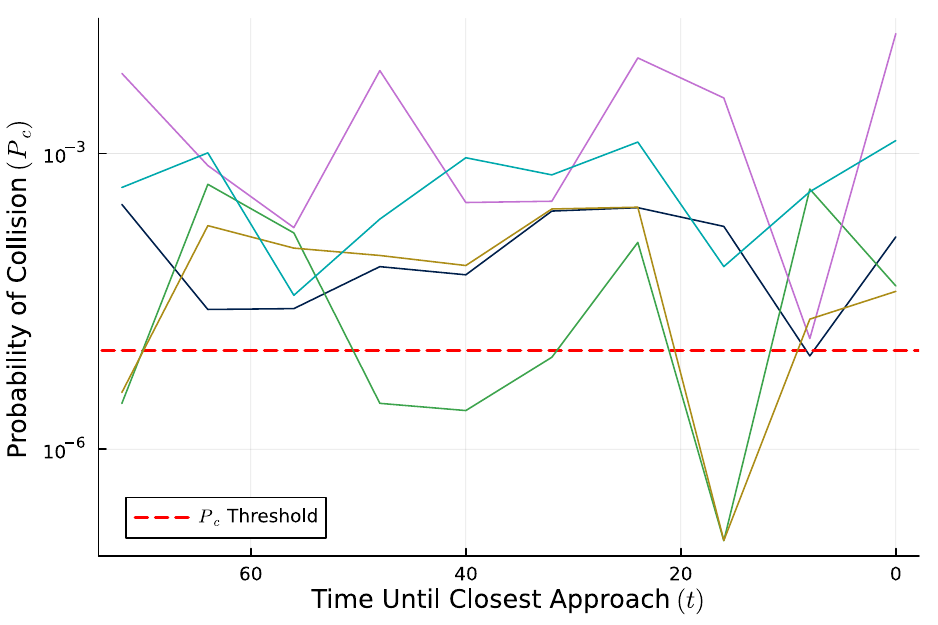}
    \caption{Examples of unsafe encounters. This plot shows how the probability of collision is updated as the time until closest approach decreases if no maneuver is taken. An unsafe encounter is defined as when the probability of collision rises above some threshold as the time until closest approach goes to 0.}
    \label{fig:unsafeenc}
\end{figure}

Our simulator generates an approximately even number of scenarios where at $t=0$ the risk is above or below some threshold $P_{c,\text{threshold}}=\num{e-5}$; this is shown in \cref{tab:sim_char}.
\begin{table}
\caption{\bf Simulated $P_c$ Distribution}
\centering
\begin{tabular}{l r}
\toprule
\bfseries $P_c$ at $t=0$ & \bfseries Proportion of Total \\
\midrule
$P_c < 1$ and $P_c < 10^{-1}$   & $2.2\times 10^{-5}$ \\
$P_c < 10^{-1}$ and $P_c < 10^{-2}$   & 0.009 \\
$P_c < 10^{-2}$ and $P_c < 10^{-3}$   & 0.262 \\
$P_c < 10^{-3}$ and $P_c < 10^{-4}$   &  0.157 \\
$P_c < 10^{-4}$ and $P_c < 10^{-5}$   & 0.056 \\
$P_c < 10^{-5}$   & 0.516  \\
\bottomrule
\end{tabular}
\label{tab:sim_char}
\end{table}
To avoid assigning an arbitrary level of risk, we categorize each encounter we generate as safe, unsafe, or trivial.
We define a safe encounter where $P_c < P_{c,\text{threshold}}$ at $t=0$ and $P_c > P_{c,\text{threshold}}$ at some $t>0$; a CDM would be issued but the risk would be mitigated by taking no action.
Examples for how $P_c$ and $R_\text{maneuver}$ progress for safe scenarios are shown in~\cref{fig:safeenc}.
An unsafe encounter is where $P_c \geq P_{c,\text{threshold}}$ at $t=0$ and $P_c > P_{c,\text{threshold}}$ at some $t>0$; a CDM would be issued and a maneuver is required to mitigate risk. 
Unsafe encounter $P_c$ and $R_\text{maneuver}$ are shown in~\cref{fig:unsafeenc}. 
A trivial encounter is defined as a scenario where $P_c \leq P_{c,\text{threshold}}$ for all $t>0$; a CDM would never have been issued so there is no decision to make.
To accurately analyze the performance of each method, performance metrics are calculated for encounter sets consisting of different percentages of safe and unsafe encounters.

\begin{figure}
    \includegraphics[width=\linewidth]{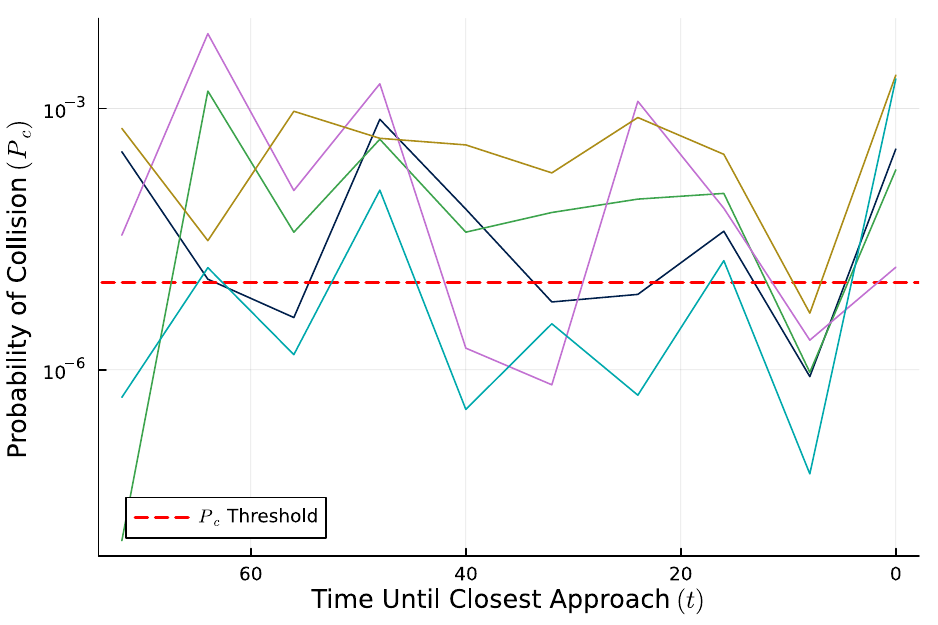}
    \caption{Examples of unsafe encounters where $P_c<P_{c,\text{threshold}}$ when $t=8$h. This is an subset of unsafe encounters in which $P_c$ rises from below the safety threshold to above it between the last state update and the time of closest approach. Consequently, not all rule-based baselines successfully mitigate risk at the final state update.}
    \label{fig:weirdenc}
\end{figure}

Because of the stochastic nature of how $P_c$ updates between states, there is a probability that at the last state update $P_c$ is below $P_{c,\text{threshold}}$ but transitions over the threshold just before $t=0$, as shown in~\cref{fig:weirdenc}.
This happens in the set of unsafe encounters with a probability of 0.06.
Consequently, the rule-based and limited horizon methods with algorithmic \enquote{guarantees} of safety still have some risk of $P_c$ increasing between $t=0$ and $t=8$ hours.
While this may be an artifact of the simulation, it explains why some methods result in unsafe encounters despite algorithmic safeguards. 
The risk this introduces could be mitigated by lowering the value of $P_{c,\text{threshold}}$ or by reducing the timestep between the last decision point and the time of closest approach.

\subsection{Results}

\begin{figure*}
    \centering
    \begin{minipage}{0.49\textwidth}
    \begin{tikzpicture}
        \begin{axis}[enlargelimits=0.2,
            grid=both, grid style={lightgray!45},
            xlabel={$\Delta v$ Required per Safe Encounter}, 
            ylabel={$\Delta v$ Required per Unsafe Encounter}, 
            title={Fuel Efficiency of All Methods},
            scale=1,
            legend pos=north west,
        ]
        \addplot[
            scatter/classes={b={red}},
            scatter, mark=*, only marks,
            scatter src=explicit symbolic,
            visualization depends on={value \thisrow{label} \as \Label},
            every node near coord/.append style={font=\small},
        ] table [meta=class] {
            x y class label
            0.01095   0.01182       b   {MCTS w/ Stochastic Depth}
            0.01112   0.01344       b   {MCTS w/ UCB1}
            0.01043   0.01546       b   {MCTS: Limited Horizon w/ Stochastic Depth}
            0.01071   0.01357       b   {MCTS: Limited Horizon w/ UCB1}
        };
        \addplot[
            scatter/classes={a={blue}, b={red}},
            scatter, mark=triangle*, only marks,
            scatter src=explicit symbolic,
            visualization depends on={value \thisrow{label} \as \Label},
            every node near coord/.append style={font=\small, anchor=south},
        ] table [meta=class] {
            x y class label
            0.01109   0.01173       a   72h
            0.01134   0.01244       a   64h
            0.01166   0.01333       a   56h
        };
        \addplot[
            scatter/classes={a={blue}, b={red}},
            scatter, mark=triangle*, only marks,
            scatter src=explicit symbolic,
            nodes near coords*={\Label},
            visualization depends on={value \thisrow{label} \as \Label},
            every node near coord/.append style={font=\small, anchor=south},
        ] table [meta=class] {
            x y class label
            0.01252   0.01539       a   40h
        };
        \addplot[
            scatter/classes={a={blue}, b={red}},
            scatter, mark=triangle*, only marks,
            scatter src=explicit symbolic,
            nodes near coords*={\Label},
            visualization depends on={value \thisrow{label} \as \Label},
            every node near coord/.append style={font=\small, anchor=north west},
        ] table [meta=class] {
            x y class label
            0.01213   0.01428       a   48h
            0.01332   0.01626       a   32h
            0.01471   0.01844       a   24h
            0.01691   0.02383       a   16h
        };
        \addplot[
            scatter/classes={a={blue}, b={red}},
            scatter, mark=triangle*, only marks,
            scatter src=explicit symbolic,
            nodes near coords*={\Label},
            visualization depends on={value \thisrow{label} \as \Label},
            every node near coord/.append style={font=\small, anchor=north},
        ] table [meta=class] {
            x y class label
            0.01842   0.03656       a   {$t_\text{cutoff} = 8$h}
        };
        \draw[dashed, gray] (axis cs:.0102,.011) rectangle (axis cs:.01185,.0165) node[pos=.0,anchor=north west] {See next plot};
        \legend{MCTS Methods,Rule-Based Baseline Methods}
        \end{axis}
        
    \end{tikzpicture}
    \end{minipage}
    \begin{minipage}{0.49\textwidth}
    \begin{tikzpicture}
        \begin{axis}[enlargelimits=0.2,
            grid=both, grid style={lightgray!45},
            xlabel={$\Delta v$ Required per Safe Encounter}, 
            ylabel={$\Delta v$ Required per Unsafe Encounter}, 
            title={Fuel Efficiency of Selected Methods},
            scale=1,
        ]
        \addplot[
            scatter/classes={b={red}},
            scatter, mark=*, only marks,
            scatter src=explicit symbolic,
            nodes near coords*={\Label},
            visualization depends on={value \thisrow{label} \as \Label},
            every node near coord/.append style={font=\small}
        ] table [meta=class] {
            x y class label
            0.01095   0.01182       b   {Full Horizon w/ Stochastic Depth}
            0.01071   0.01357       b   {Limited Horizon w/ UCB1}
        };
        \addplot[
            scatter/classes={b={red}},
            scatter, mark=*, only marks,
            scatter src=explicit symbolic,
            nodes near coords*={\Label},
            visualization depends on={value \thisrow{label} \as \Label},
            every node near coord/.append style={font=\small, anchor = west}
        ] table [meta=class] {
            x y class label
            0.01043   0.01546       b   {Limited Horizon w/ Stochastic Depth}
        };
        \addplot[
            scatter/classes={b={red}},
            scatter, mark=*, only marks,
            scatter src=explicit symbolic,
            nodes near coords*={\Label},
            visualization depends on={value \thisrow{label} \as \Label},
            every node near coord/.append style={font=\small, anchor=north}
        ] table [meta=class] {
            x y class label
            0.01112   0.01344       b   {Full Horizon w/ UCB1}
        };
        \addplot[
            scatter/classes={a={blue}},
            scatter, mark=triangle*, only marks,
            scatter src=explicit symbolic,
            nodes near coords*={\Label},
            visualization depends on={value \thisrow{label} \as \Label},
            every node near coord/.append style={font=\small, anchor = north west}
        ] table [meta=class] {
            x y class label
            0.01109   0.01173       a   72h
            0.01166   0.01333       a   56h
        };
        \addplot[
            scatter/classes={a={blue}},
            scatter, mark=triangle*, only marks,
            scatter src=explicit symbolic,
            nodes near coords*={\Label},
            visualization depends on={value \thisrow{label} \as \Label},
            every node near coord/.append style={font=\small, anchor = south}
        ] table [meta=class] {
            x y class label
            0.01134   0.01244       a   64h
        };
        \end{axis}
    \end{tikzpicture}
    \end{minipage}
    \caption{Each maneuvering algorithm is plotted with rule-based methods in blue and MCTS methods in red. The horizontal axis represents the average $\Delta v$ spent per safe encounter, while the vertical axis is the average $\Delta v$ spent per unsafe encounter. The rule-based method with $t_\text{cutoff}=72$ hours is the theoretical lower bound on $\Delta v$ for unsafe encounters.}
    \label{fig:AllScatterPlot}
\end{figure*}
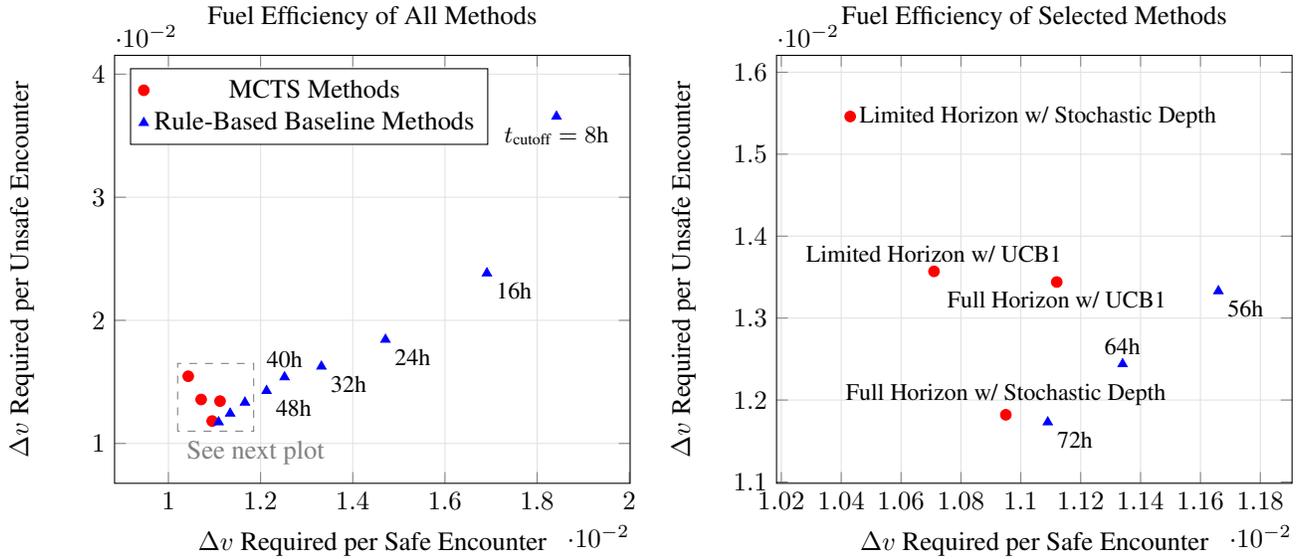

We generated 90,000 scenarios and compared the average cost for mitigating risk between policies.
The results for sets of completely safe and unsafe encounters are presented in \cref{results}.  
\begin{table*}
\centering
\caption{\bf Performance Metrics}
\begin{tabular}{l r r r}
\toprule
\bfseries Maneuver Planner & \bfseries Cost per Safe Encounter $\Delta v$ & \bfseries Cost per Unsafe Encounter $\Delta v$ & \bfseries $p_\text{success}$ \\
\midrule
MCTS w/ Full Horizon W/ Stochastic Depth & 0.01095 & 0.01182 & 0.999\\
MCTS w/ Full Horizon W/ UCB1   & 0.01112 & 0.01344 & 0.998  \\
MCTS w/ Limited Horizon w/ Stochastic Depth & 0.01043 & 0.01546 & 0.982  \\
MCTS w/ Limited Horizon w/ UCB1   & 0.01071 & 0.01357 & 0.994  \\
Rule-Based: $t_\text{cutoff}=72$h & 0.01109 & 0.01173 & 1.0 \\
Rule-Based: $t_\text{cutoff}=64$h &  0.01134 & 0.01244 & 0.999\\
Rule-Based: $t_\text{cutoff}=56$h & 0.01166 & 0.01333 & 0.999\\
Rule-Based: $t_\text{cutoff}=48$h & 0.01213 & 0.01428 & 0.998\\
Rule-Based: $t_\text{cutoff}=40$h & 0.01252 & 0.01539 & 0.996\\
Rule-Based: $t_\text{cutoff}=32$h & 0.01332 & 0.01626 & 0.993\\
Rule-Based: $t_\text{cutoff}=24$h & 0.01471 & 0.01844 & 0.989\\
Rule-Based: $t_\text{cutoff}=16$h & 0.01691 & 0.02383 & 0.978\\
Rule-Based: $t_\text{cutoff}=8$h  & 0.01842 & 0.03656 & 0.939\\
\bottomrule
\end{tabular}
\label{results}
\end{table*}
\cref{fig:AllScatterPlot} visualizes these results. For unsafe encounters, we can treat the rule-based method where $t_\text{cutoff}=72$ as the theoretical lower bound on fuel expended per encounter.
If we had an agent that automatically knew that an encounter would be unsafe but not how the uncertainty would progress, the optimal choice would be to maneuver at the first CDM.
For safe encounters, the lower bound on $\Delta v$ is 0.

Each method successfully mitigates unsafe encounters with probability $p_\text{success}$. 
These results are shown in~\cref{results}.
We treat $p_\text{success}$ at the last information update when $t=8$ hours as the standard for safety. 
By definition $p_\text{success}$ is 1.0 when $t=72$ hours as encounters are only tested when a CDM has been issued at some point where $t>0$.

\subsubsection{Rule-Based Policies}
\cref{fig:AllScatterPlot} shows all rule-based methods as blue data points. 
\cref{results} describes which point is associated with each method.
The method with $t_\text{cutoff}=8$ expends the most fuel on average for any encounter and $t_\text{cutoff}=72$ expends less fuel on average compared with any rule-based policy.
The average cost per maneuver is linearly correlated between these two as $t_\text{cutoff}$ is varied. 
This shows that given the constraints of only along-track and anti-along-track maneuvers, the value of future information in this simulation is outweighed by the fuel needed when a maneuver is required. 
It is important to note that this conclusion is specific to this simulation and the assumptions made in how the uncertainty over $x_c$ and $x_d$ update over time.

\subsubsection{MCTS}
The MCTS implementations used the hyperparameters in~\cref{hyperparameters}.
\begin{table}
\caption{\bf MCTS Hyperparameters}
\centering
\begin{tabular}{l r}
\toprule
\bfseries Parameter & \bfseries Value \\
\midrule
$p_\text{obs}$   &  $0.5$  \\
$k_{d,\text{lower}}$   & $0.05$ \\
$k_{d,\text{upper}}$ & $0.3$\\
$k_c$   & $0.05$ \\
$R_{crash}$   &  $-0.5$ \\
$N_{sim}^{max}$   & $200$ \\
$\gamma$   &  $0.95$ \\
$c$   & $0.8$ \\
$t_\text{maxdepth}$ & 8h \\
\bottomrule
\end{tabular}
\label{hyperparameters}
\end{table}
\cref{fig:AllScatterPlot} shows the average cost per encounter for the MCTS methods compared with select rule-based methods.
Unlike with the rule-based methods, there is not a linear relationship between the fuel used by each MCTS method.
MCTS with the stochastic depth heuristic and a full horizon uses the least amount of fuel for unsafe encounters, approaching  the lower bound of the rule-based policy when $t_\text{cutoff} = 72$ hours.
However, while it uses less fuel than all rule based policies per safe encounter, it uses more fuel than both MCTS methods with a limited horizon. 
This performance comes with the second best proportion of encounters where risk is mitigated $p_\text{success}$, second only to the rule-based policy with $t_\text{cutoff} = 72$ hours where perfect safety is guaranteed. 
MCTS with a full horizon and the UCB1 heuristic uses more fuel in both safe and unsafe encounters and assumes slightly more risk.

When a limited horizon is applied, fuel expended per safe encounter decreases; however, this comes with a cost of increased fuel consumption for unsafe encounters and risk assumed. 
This is more pronounced with the stochastic depth heuristic than with the UCB1 heuristic. 
Both limited horizon implementations have safety margins better than standard set by the rule-based method with $t_\text{cutoff}=8$.

\subsubsection{Optimal Methods}

The net fuel cost for a satellite is determined by the sum of safe and unsafe encounters it experiences.
Because an operator sets a method over an extended period of time, the estimated proportion of safe to unsafe encounters $p_\text{safe}$ determines the optimal policy.
By linearly combining the cost per maneuver for safe and unsafe encounters and comparing them, we are able to estimate the optimal policy for different values of $p_\text{safe}$.
This is calculated by minimizing the estimated maneuver cost function
\begin{equation}
    R_{\pi,\text{est}}(p_\text{safe}) = p_\text{safe}R_{\pi,\text{safe}}+(1-p_\text{safe})R_{\pi,\text{unsafe}}
\end{equation}
where $R_{\pi,\text{safe}}$ is the average reward per safe encounter and $R_{\pi,\text{unsafe}}$ is the average reward per unsafe encounter for method $\pi$.

When $p_\text{safe} > 0.874$, MCTS with full horizon and the stochastic depth heuristic was the optimal choice. 
If $p_\text{safe}<0.378$, the optimal solution is the rule-based policy when $P_{c,\text{threshold}}=72$ hours.
Otherwise, MCTS with limited horizon and using the stochastic depth heuristic had the lowest cost. 

\section{Conclusion}

This paper introduced a formulation for the satellite collision avoidance problem as a Markov decision process. 
The MDP was solved with Monte Carlo methods and compared against baseline static rule-based methods.
We introduced two variations of Monte Carlo tree search with two different exploration heuristics. 
The results show that the optimal method for a set of encounters depends on the distribution of the probability of collision in that encounter set.
We observed that Monte Carlo methods outperformed the current rule-based baseline methods for all cases when less than 62.2\% of close encounters result in a collision.
MCTS methods reduced both the average fuel required per encounter and the amount of risk assumed by the agent compared to rule-based policies. 
Future work would benefit from relaxing some of the constraints imposed on this problem. 
Specifically, allowing maneuvers other than along-track and anti-along-track might enable more fuel-efficient maneuvers.
Another expansion would be to broaden the action space to include incremental maneuvers so the satellite could take small incremental actions to mitigate risk, instead of a binary choice of maneuvering or waiting. 

\appendices{}              
\crefalias{section}{appendix}

\section{Calculating the Cost to Maneuver}        
\label{appendix:cost2maneuver}

We calculate the cost to maneuver by iterative calculating how much $\Delta v$ is required to reduce the probability of collision at a state $P_c(s)$ below a certain risk threshold $P_{c,\text{threshold}}$. 
The chief state $x_c$ is backpropagated to $t$ using Keplerian dynamics.
This backpropagated chief state $\bar x_c$ is broken into its ECI position $u$ and velocity $v$.
Then, one unit of velocity change $\delta_{\Delta v}$ is added to the velocity $v_c$ in both the along track and anti-along track direction.
Both cases are then propagated back forward to $t=0$ and a new $P_c$ is calculated. 
This process is repeated in whichever direction made a greater improvement in $P_c$. 
Once $P_c<P_{c,\text{threshold}}$, the total number of times $\delta_{\Delta v}$ was added in a direction is summed and returned as the total cost to maneuver.
This process is outlined in \cref{alg:costtomaneuver}.
\begin{algorithm}
    \caption{Cost to Maneuver}
    \begin{algorithmic}[1] 
        \Function{ApplyThrust}{$x_c$, $t$,Direction}
            \State $\bar x_c \leftarrow$ \Call{BackPropagate}{$x_c$,$t$}
            \State $u,v \leftarrow \bar x_c$
            \If {Direction $=$ along\_track}
                \State $v' \leftarrow \frac{v}{|v|} \times {|v| \delta_{\Delta v}} $
            \Else
                \State $v' \leftarrow \frac{v}{|v|} \times {|v| \delta_{\Delta v}} $
            \EndIf
            \State $\bar x_c' \leftarrow u,v'$
            \State $x_c' \leftarrow$ \Call{ForwardPropagate}{$\bar x_c,t$}
            \State \Return $x_c'$
        \EndFunction
        \Function {CosttoManeuver} {$s$}
            \State \textbf{assert:} $P_c(s) > P_{c,\text{threshold}}$ 
            \State $t, x_c,x_d, \Sigma_c, \Sigma_d, r_c, r_d \leftarrow s$
            \State $x_{c,AT}' \leftarrow$ \Call{ApplyThrust}{$x_c$, $t$, along\_track}
            \State $x_{c,AAT}' \leftarrow$ \Call{ApplyThrust}{$x_c$, $t$, anti\_along\_track}
            \State $s_{AT}' \leftarrow (t, x_{c,AT}', x_d, \Sigma_c, \Sigma_d, r_c, r_d)$
            \State $s_{AAT}' \leftarrow (t, x_{c,AAT}', x_d, \Sigma_c, \Sigma_d, r_c, r_d)$
            \If{$P_c(s_{AT}') > P_c(s_{AAT}')$}
                \State ThrustDirection $\leftarrow$ anti\_along\_track
                \State $P_{c,current} \leftarrow P_c(s_{AAT}')$
                \State $s_{c,current} \leftarrow s_{AAT}'$
            \Else
                \State ThrustDirection $\leftarrow$ along\_track
                \State $P_{c,current} \leftarrow P_c(s_{AT}')$
                \State $s_{c,current} \leftarrow s_{AT}'$
            \EndIf
            \State $i \leftarrow 1$
            \State total\_cost $\leftarrow \delta_{\Delta v}$
            \While{$P_{c,current} \geq P_{c,\text{threshold}}$}
                \State $t, x_c, x_d, \Sigma_c, \Sigma_d, r_c, r_d \leftarrow s_{c,current}$
                \State $x_c' \leftarrow$ \Call{ApplyThrust}{$x_c$, $t$ ThrustDirection}
                \State $s_{c,current}' \leftarrow t, x_c',x_d,\Sigma_c,\Sigma_d,r_c,r_d$
                \State $P_{c,current} \leftarrow P_c(s')$
                \State $s_{c,current} \leftarrow s_{c,current}'$
                \State $i \leftarrow i + 1$
                \State total\_cost $\leftarrow \text{total\_cost}+\delta_{\Delta v}$
            \EndWhile
            \State \Return total\_cost
        \EndFunction
    \end{algorithmic}
    \label{alg:costtomaneuver}
\end{algorithm}

\section{Generating Collision Scenarios}
\label{appendix:scenario_gen}
This appendix outlines the methodology used for calculating the deputy initial position and velocity $x_{d,0}$ and how the simulated states $s_{c,\text{current}}$ are generated.

\subsection{Deputy State}
Given the initial chief position and velocity $x_{c,0}$, we sample the deputy position and velocity $x_{d,0}$. 
We assign the initial deputy position $u_{d,0}$ by sampling from a normal distribution ${\mathcal N}(\mu_{u_{c,0}},\Sigma_{u_{c,0}})$ over the initial chief position $u_{c,0}$. 
We assume that the magnitude of the deputy velocity $v_{d,0}$ is the same as the magnitude of the chief velocity $v_{c,0}$. 
The direction of the deputy velocity must be perpendicular to the vector from the spacecraft to the earth; the velocity direction must also be perpendicular to the vector from the deputy to the chief.
This forces the deputy to be close to its position at the time of closest approach. 
This is outlined in \cref{alg:deputystate}.
\begin{algorithm}
    \caption{Deputy State Generation}
    \begin{algorithmic}[1] 
        \Function{Generate Deputy}{}
        \State $u_{c,0},v_{c,0} \leftarrow x_{c,0}$
        \State $u_{d,0} \sim {\mathcal N}(\mu_{u_{c,0}},\Sigma_{u_{c,0}})$
        \State $r_{c,d} \leftarrow u_{d,0} - u_{c,0}$
        \State $A \leftarrow \begin{bmatrix} u_{d,0} \\ r_{c,d} \\ (0,0,1) \end{bmatrix}$
        \State $z \sim {\mathcal U}(-1,1)$
        \State $b \leftarrow \begin{bmatrix} 0 & 0 & z\end{bmatrix}$
        \State ${v_{d,0}}\leftarrow (A \backslash b) \times |v_{c,0}|$
        \State $x_{d,0} \leftarrow u_{d,0},v_{d,0}$
        \State \Return $x_{d,0}$
        \EndFunction
    \end{algorithmic}
    \label{alg:deputystate}
\end{algorithm}

\subsection{Covariance Propagation, State Sampling, and Encounter Generation}

We initialize uncertainty covariance matrices $\Sigma_{c,0}$ and $\Sigma_{d,0}$ at the time of closest approach as diagonal matrices with diagonal values sampled the normal distribution defined by ${\mathcal N}(\mu_{\Sigma_c},\Sigma_{\Sigma_c})$ and ${\mathcal N}(\mu_{\Sigma_d},\Sigma_{\Sigma_d})$.
The means for the distributions $\mu_{\Sigma_d}$ and $\mu_{\Sigma_c}$ had the same order of magnitude as the example described in the \textit{CMD Summary Message Guide}~\cite{noauthor_conjunction_nodate}. 
These matrices $\Sigma_{c,0}$ and $\Sigma_{d,0}$ are then backpropagated based on the expected measurement characteristics of the spacecraft.
The chief covariance matrix $\Sigma_c$ is scaled by the factor $k_{c,\text{backprop}}$ every timestep so that the farther it is from the time of closest approach, the larger the covariance will scale. The deputy covariance matrix $\Sigma_d$ is scaled by the factor $k_{d,\text{backprop}}$ sampled from the distribution $\mathcal U$$(k_{d,\text{lower}},k_{d,\text{upper}})$. To simulate infrequent measurements, this update happens with a probability $p_\text{obs}$, otherwise no update happens. 
A chief and deputy state is sampled from each of the covariances generated to make synthetic state \enquote{measurements} that are then used to evaluate the models. 
Each new state is added to the encounter set $S$, which is returned.
This process is outlined in \cref{alg:statesampling}.

\begin{algorithm}
    \caption{Generate Encounter}
    \begin{algorithmic}[1] 
        \Function{GenerateEncounter}{$x_{c,0},x_{d,0}$}
        \State $r_{c} \sim {\mathcal U}(5,10)$
        \State $r_{d} \sim {\mathcal U}(5,10)$
        \State $\vec \sigma_{c,0} \sim {\mathcal N}(\mu_{\Sigma_c},\Sigma_{\Sigma_c})$
        \State $\Sigma_{c,0} \leftarrow \text{diag}(\vec \sigma_{c,0})$
        \State $\vec \sigma_{d,0} \sim {\mathcal N}(\mu_{\Sigma_d},\Sigma_{\Sigma_d})$
        \State $\Sigma_{d,0} \leftarrow \text{diag}(\vec \sigma_{d,0})$
        \State $\hat x_{c,0} \sim {\mathcal N}(x_0,\Sigma_{c,0})$
        \State $\hat x_{d,0} \sim {\mathcal N}(x_0,\Sigma_{c,0})$
        \State $t_0 \leftarrow 0$
        \State $s_0 \leftarrow (t_0,x_{c,0},x_{d,0},\Sigma_{c,0},\Sigma_{d,0},r_{b,hb},r_{d})$
        \State $S \leftarrow s_0$
        \State $i \leftarrow 8$
        \While{$i<72$}
            \State $t \leftarrow i$
            \State $\Sigma_{c,i} \leftarrow \Sigma_{c,i-8}(1+k_c)$
            \State $\text{is\_observed} \sim {\mathcal{U}}(0,1)$
            \If{$\text{is\_observed}<p_\text{obs}$}
                \State $k_d \sim {\mathcal U}(k_{d,lower},k_{d,upper})$
            \Else
                \State $k_d \leftarrow 0$
            \EndIf
            \State $\Sigma_{d,i} \leftarrow \Sigma_{d,i-8}(1+k_d)$
            \State $\hat x_{c,i} \sim {\mathcal{N}}(x_{c,0},\Sigma_{c,i})$
            \State $\hat x_{d,i} \sim {\mathcal{N}}(x_{d,0},\Sigma_{d,i})$
            \State $s_i \leftarrow (t, \hat x_{c,i}, \hat x_{d,i}, \Sigma_{c,i}, \Sigma_{d,i}, r_{c}, r_{d})$
            \State $S \leftarrow S \cup s_i$
            \State $i \leftarrow i+8$
        \EndWhile
        \State \Return $S$
        \EndFunction
    \end{algorithmic}
    \label{alg:statesampling}
\end{algorithm}


\newcommand{\newblock}{\ }
\bibliographystyle{IEEEtranN}
\bibliography{ref.bib}

\thebiography

\begin{biographywithpic}
{William (Bill) Kuhl}{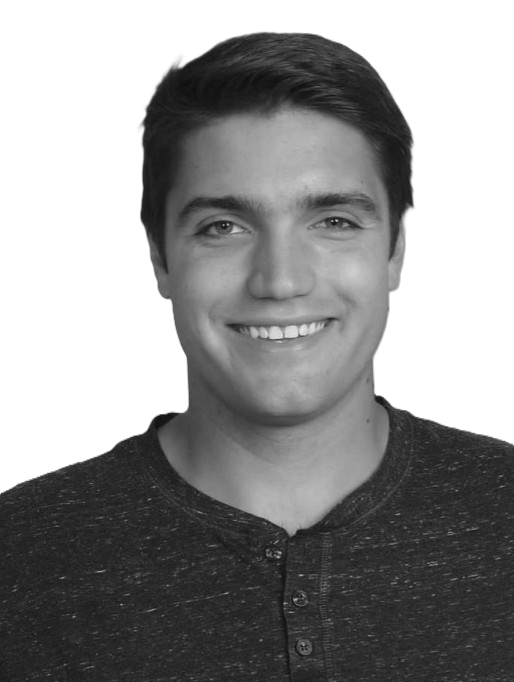} is a 2\textsuperscript{nd} Lieutenant in the US Space Force and graduate researcher in the Stanford Intelligent Systems Laboratory (SISL).
His research focuses on decision-making in space systems.
He recieved his B.S. in Aerospace Engineering from the Massachussetts Institute of Technology in 2022, and a masters degree in Global Affairs from Tsinghua University in 2023.
\end{biographywithpic}

\begin{biographywithpic}
{Jun Wang}{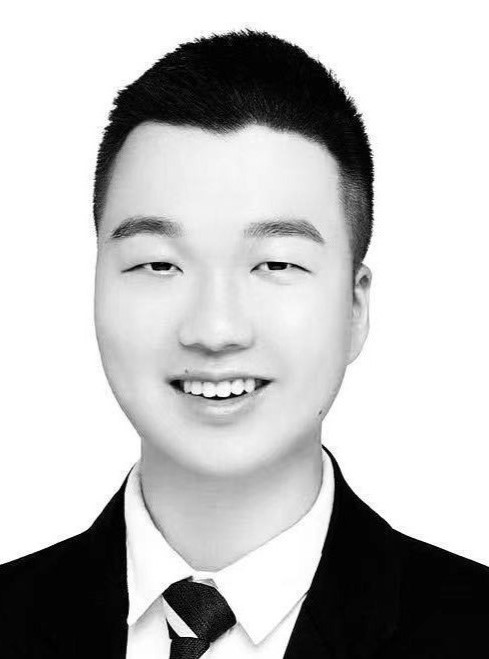} is an undergraduate Computer Science and Mathematics student at Stanford 
University who is currently a research assistant at the Stanford Intelligent Systems Laboratory 
(SISL). His undergraduate academic concentration is Artificial Intelligence, and his research 
interests include autonomous decision-making, deep generative models, and bioinformatics. He 
is especially interested in the space applications of decision-making algorithms and has been 
involved in the Stanford Student Space Initiative (Stanford SSI)
\end{biographywithpic}
\begin{biographywithpic}{Duncan Eddy}{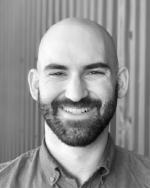}
is a postdoctoral research fellow in the Stanford Intelligent Systems Laboratory (SISL) and the Executive Director of the Center for AI Safety at Stanford University. His research focuses on decision-making in safety-critical, climate, and space systems. He received B.S. in Mechanical Engineer from Rice University in 2013, and PhD in Aerospace Engineering from Stanford University in 2021. Prior to returning to Stanford he was the Director of Space Operations at Capella Space Corporation where he built a fully automated constellation tasking and delivery system. He later founded and led the Constellation Management and Space Safety Organization at Amazon's Project Kuiper.
\end{biographywithpic}

\begin{biographywithpic}{Mykel Kochenderfer}{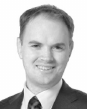}
 is Associate Professor of Aeronautics and Astronautics and Associate Professor, by courtesy, of Computer Science at Stanford University. He is the director of the Stanford Intelligent Systems Laboratory (SISL), conducting research on advanced algorithms and analytical methods for the design of robust decision-making systems. Prior to joining the faculty in 2013, he was at MIT Lincoln Laboratory where he worked on airspace modeling and aircraft collision avoidance. He received his Ph.D. from the University of Edinburgh in 2006 where he studied at the Institute of Perception, Action and Behaviour in the School of Informatics. He received B.S. and M.S. degrees in computer science from Stanford University in 2003.
\end{biographywithpic}

\end{document}